\pdfoutput=1

\documentclass[11pt]{article}

\usepackage{naacl2021}

\usepackage{times}
\usepackage{latexsym}
\usepackage{multirow}
\usepackage{tabularx}
\usepackage{amsmath}
\usepackage{amssymb}
\usepackage{bm}
\usepackage{subfigure}
\usepackage{latexsym}
\usepackage{arydshln}
\usepackage{microtype}
\usepackage[switch]{lineno}
\usepackage{helvet} 
\usepackage{courier}  
\usepackage{graphicx} 
\usepackage{natbib}  
\usepackage{caption} 
\usepackage[T1]{fontenc}

\usepackage[utf8]{inputenc}

\usepackage{microtype}
\title{Pruning-then-Expanding Model for Domain Adaptation \\ of Neural Machine Translation}

\author{Shuhao Gu\textsuperscript{\rm 1,2}, Yang Feng\textsuperscript{\rm 1,2}\thanks{Corresponding author: Yang Feng. \newline \indent ~~ Reproducible code: https://github.com/ictnlp/PTE-NMT. }, \textbf{Wanying Xie\textsuperscript{\rm 1,3}}\\ 
\textsuperscript{\rm 1} Key Laboratory of Intelligent Information Processing,\\ Institute of Computing Technology, Chinese Academy of Sciences (ICT/CAS)\\
\textsuperscript{\rm 2} University of Chinese Academy of Sciences\\
\textsuperscript{\rm 3} Beijing Language and Culture University, China\\
{ \{gushuhao19b,fengyang\}@ict.ac.cn, \indent xiewanying07@gmail.com}
}

\begin{document}
\maketitle
\begin{abstract}
Domain Adaptation is widely used in practical applications of neural machine translation, which aims to achieve good performance on both general domain and in-domain data. However, the existing methods for domain adaptation usually suffer from catastrophic forgetting, large domain divergence, and model explosion. To address these three problems, we propose a method of ``divide and conquer'' which is based on the importance of neurons or parameters for the translation model. In this method, we first prune the model and only keep the important neurons or parameters, making them responsible for both general-domain and in-domain translation. Then we further train the pruned model supervised by the original whole model with knowledge distillation. Last we expand the model to the original size and fine-tune the added parameters for the in-domain translation. We conducted experiments on different language pairs and domains and the results show that our method can achieve significant improvements compared with several strong baselines. 
\end{abstract}

\section{Introduction}

Neural machine translation (NMT) models~\cite{kalchbrenner2013recurrent,ChoMGBBSB14,sutskever2014sequence,BahdanauCB14,GehringAGYD17,VaswaniSPUJGKP17} are data-driven and hence require large-scale training data to achieve good performance~\cite{zhang2019bridging}. In practical applications, NMT models usually need to produce translation for some specific domains with only a small quantity of in-domain data available, so domain adaptation is applied to address the problem. A typical domain adaptation scenario as discussed in~\citet{FreitagA16} is that an NMT model have been trained with large-scale general-domain data and then is adapted to specific domains, hoping the model can fit in-domain data well meanwhile the performance will not degrade too much on the general domain. 

Towards this end, many researchers have made their attempts. The fine-tuning method~\cite{luong2015stanford} performs in-domain training based on the general-domain model by first training the model on general-domain data and then continuing to train on in-domain data. Despite its convenience for use and high-quality for in-domain translation, this method suffers from catastrophic forgetting which leads to poor performance in the previous domains.
Regularization-based methods~\cite{dakwale2017fine,ThompsonGKDK19,BaroneHGS17,KhayrallahTDK18} instead introduce an additional loss to the original objective so that the translation model can trade off between general-domain and in-domain. This kind of methods usually has all the parameters shared by general-domain and in-domain, with the assumption that the optimal parameter spaces for all the domains will overlap with each other, and retaining these overlapped parameters can balance over all the domains. This assumption is feasible when the domains are similar,  but when the divergence of the domains is large, it is not reasonable anymore. In contrast, the methods with domain-specific networks~\cite{dakwale2017fine,abs-1911-09912,BapnaF19,GuFL19} can be often (but not always) immune to domain divergence as it can capture domain-specific features. But unfortunately, as the number of domains increases, the parameters of this kind of methods will surge. Besides, the structure of these networks needs to be carefully designed and tuned, which prevents them from being used in many cases. 

Given the above, we propose a method of domain adaptation that can not only deal with large domain divergence during domain transferring but also keep a stable model size even with multiple domains. Inspired by the analysis work on NMT~\cite{BauBSDDG19,VoitaTMST19,gu2020investigating}, we find that only some important parameters in a well-trained NMT model play an important role when generating the translation and unimportant parameters can be erased without affecting the translation quality too much. 
According to these findings, we can preserve important parameters for general-domain translation, while tuning unimportant parameters for in-domain translation. 
To achieve this, we first train a model on the general domain and then shrink the model with neuron pruning or weight pruning methods, only retaining the important neurons/parameters. To ensure the model can still perform well on general-domain data,
we adjust the model on in-domain data with knowledge distillation where the original whole model is used as the teacher and the pruned model as the student. Finally, we expand the model to the original size and fine-tune the added parameters on the in-domain data.
Experimental results on different languages and domains show that our method can avoid catastrophic forgetting on general-domain data and achieve significant improvements over strong baselines on multiple in-domain data sets. 

Our contributions can be summarized as follows:
\begin{itemize}
  \item We prove that the parameters that are unimportant for general-domain data can be utilized to improve in-domain translation quality.
  \item Our model can keep superior performance over baselines even when continually transferring to multiple domains.
  \item Our model can fit in the continual learning scenario where the data for the previous domains cannot be got anymore which is the common situation in practice.
\end{itemize}

\section{Background}
\subsection{The Transformer}
In our work, we apply our method in the framework of \textsc{Transformer}~\cite{VaswaniSPUJGKP17} which will be briefly introduced here. 
However, we note that our method can also be combined with other NMT architectures. 
We denote the input sequence of symbols as $\mathbf{x}=(x_1,\ldots,x_J)$, the ground-truth sequence as $\mathbf{y}^{*}=(y_1^{*},\ldots,y_{K*}^{*})$ and the translation as $\mathbf{y}=(y_1,\ldots,y_K)$.

\noindent \textbf{The Encoder \& Decoder} The encoder is composed of $\mathnormal{N}$ identical layers. Each layer has two sublayers. The first is a multi-head self-attention sublayer and the second is a fully connected feed-forward network. Both of the sublayers are followed by a residual connection operation and a layer normalization operation. The input sequence $\mathbf{x}$ will be first converted to a sequence of vectors $\mathbf{E}_x=[E_x[x_1];\ldots;E_x[x_J]]$ where $E_x[x_j]$ is the sum of word embedding and position embedding of the source word $x_j$. 
Then, this sequence of vectors will be fed into the encoder and the output of the $\mathnormal{N}$-th layer will be taken as source hidden states. and we denote it as $\mathbf{H}$. 
The decoder is also composed of $\mathnormal{N}$ identical layers. In addition to the same kind of two sublayers in each encoder layer, the cross-attention sublayer is inserted between them, which performs multi-head attention over the output of the encoder. The final output of the $\mathnormal{N}$-th layer gives the target hidden states $\mathbf{S}=[\mathbf{s}_1;\ldots;\mathbf{s}_{K*}]$, where $\mathbf{s}_k$ is the hidden states of $y_k$. 

\noindent \textbf{The Objective} 
We can get the predicted probability of the $k$-th target word over the target vocabulary by performing a linear transformation and a softmax operation to the target hidden states:
\begin{equation}
    p(y_k | \mathbf{y}_{<k}, \mathbf{x}) \propto \exp({\mathbf W}_o  {\mathbf s}_k + \mathbf{b}_o),
\end{equation}
where ${\mathbf W}_o \in \mathbb{R}^{d_{model}\times|\mathrm{V}_t|}$ and $|\mathrm{V}_t|$ are the size of target vocabulary. 
The model is optimized by minimizing a cross-entropy loss of the ground-truth sequence with teacher forcing training:
\begin{equation}\label{eq::loss}
    \mathcal{L}(\theta) = -\frac{1}{K} \sum_{k=1}^{K} \log p(y_k^{*} | \mathbf{y}_{<k}, \mathbf{x}; \theta),
\end{equation}
where $K$ is the length of the target sentence and $\theta$ denotes the model parameters. 

\subsection{Knowledge Distillation}
Knowledge Distillation (KD) method~\cite{HintonVD15} is for distilling knowledge from a teacher network to a student network. Normally, the teacher network is considered to be with higher capability.
A smaller student network can be trained to perform comparablely or even better by mimicking the output distribution of the teacher network on the same data. This is usually done by minimizing the cross entropy between the two distributions:
\begin{equation}
\begin{split}
    \mathcal{L}_{\mathrm{KD}}(\theta, \theta_T) = -\frac{1}{K} & \sum_{k=1}^{K} q(\mathbf{y}_k | \mathbf{y}_{<k}, \mathbf{x}; \theta_T) \\
     & \times \log p(\mathbf{y}_k| \mathbf{y}_{<k}, \mathbf{x}; \theta),
\end{split}
\end{equation}
where $q$ denotes the output distribution of the teacher network and $\theta$ and $\theta_T$ denote the parameters of the student and teacher network, respectively. 
The parameters of the teacher network usually keep fixed during the KD process.

\begin{figure*}[t!]
    \centering
    \includegraphics[width=2.0\columnwidth]{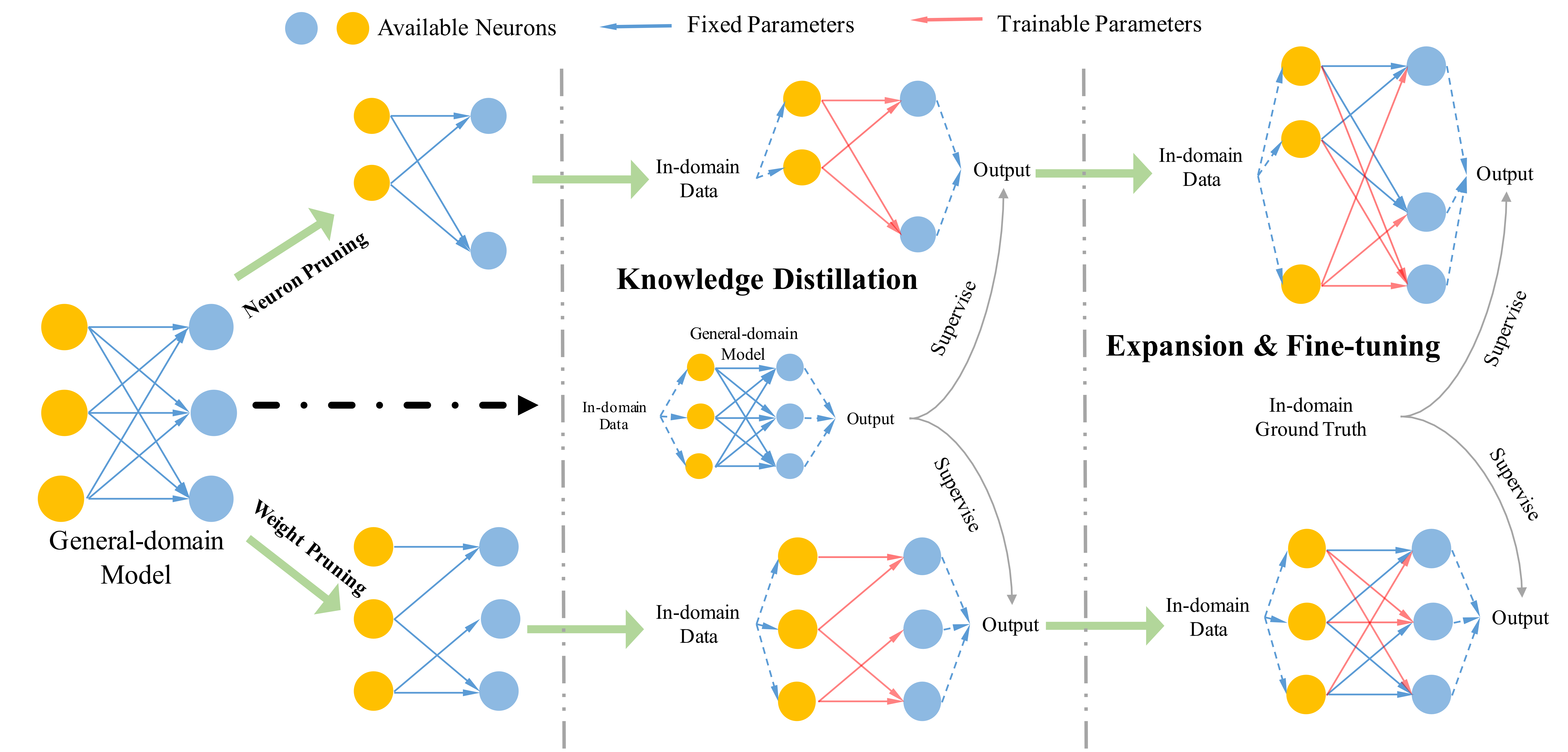}
    \caption{The whole training process of the proposed method.}
    \label{fig:method}
\end{figure*}

\section{Method}\label{method}
The main idea of our method is that different neurons or parameters have different importance to the translation model and hence different roles in domain adaptation. Based on this, we distinguish them into important and unimportant ones and make important neurons or parameters compromise between domains while unimportant ones focus on in-domain. Specifically, our method involves the following steps shown in Figure~\ref{fig:method}. First, we train a model on the general domain and then evaluate the importance of different neurons or parameters. Then we erase the unimportant neurons or parameters and only keep the ones that are related to the general domain so that our method will not be subjected to domain divergence. Next, we further adjust our model under the framework of knowledge distillation~\cite{HintonVD15} on the in-domain with the unpruned model as the teacher and the pruned model as the student. In this way, the pruned model can regain some of its lost performance because of pruning. Finally, we expand the pruned model to the original size and fine-tune the added parameters for the in-domain.

\subsection{Model Pruning}
Model pruning aims to find a good subset of neurons and parameters of the general-domain model while maintaining the original performance as much as possible. 
Therefore, under the premise of retaining most of the model's capability, we want to remove those unimportant neurons or parameters to reduce the size of the whole model first. To achieve this, we adopt two pruning schemes. The first is neuron pruning, where we evaluate the importance of neurons directly and then prune unimportant neurons and relevant parameters. The second is weight pruning, where we evaluate and prune each parameter directly.

\textbf{Neuron Pruning} To evaluate the importance of each neuron, we adopt a criterion based on the Taylor expansion~\cite{MolchanovTKAK17}, where we directly approximate the change in loss when removing a particular neuron. Let $h_i$ be the output produced from neuron $i$ and $H$ represents the set of other neurons. 
Assuming the independence of each neuron in the model, the change of loss when removing a certain neuron can be represented as:
\begin{equation}
    |\Delta\mathcal{L}(h_i)| = |\mathcal{L}(H, h_i=0) - \mathcal{L}(H, h_i)|,
\end{equation} 
where $\mathcal{L}(H, h_i=0)$ is the loss value if the neuron $i$ is pruned and $\mathcal{L}(H, h_i)$ is the loss if it is not pruned. For the function $\mathcal{L}(H, h_i)$, its Taylor expansion at point $h_i = a$ is:
\begin{equation}
  \mathcal{L}(H, h_i) = \sum_{n=0}^{N}\frac{\mathcal{L}^{n}(H, a)}{n!}(h_i - a)^n + R_N(h_i),
\end{equation}
where $\mathcal{L}^{n}(H, a)$ is the $n$-th derivative of $\mathcal{L}(H, h_i)$ evaluated at point $a$ and $R_N(h_i)$ is $N$-th remainder. 
Then, approximating $\mathcal{L}(H, h_i=0)$ with a first-order Taylor polynomial where $h_i$ equals zero:
\begin{equation}
    \mathcal{L}(H, h_i=0) = \mathcal{L}(H, h_i)-\frac{\partial \mathcal{L}(H, h_i)}{\partial h_i}h_i-R_1(h_i).
\end{equation}
The remainder $R_1$ can be represented in the form of Lagrange:
\begin{equation}
    R_1(h_i) = \frac{\partial^2\mathcal{L}(H, h_i)}{\partial^2\delta h_i}h_i^2,
\end{equation}
where $\delta \in (0,1)$. Considering the use of ReLU activation function in the model, the first derivative of loss function tends to be constant, so the second order term tends to be zero in the end of training. Thus, we can ignore the remainder and get the importance evaluation function as follows:
\begin{equation}
    \Theta_\mathrm{TE}(h_i) = \left|\Delta\mathcal{L}(h_i)\right|  =  \left|\frac{\partial \mathcal{L}(H, h_i)}{\partial h_i}h_i\right|.
\end{equation} 
In practice, we need to accumulate the product of the activation and the gradient of the objective function w.r.t to the activation, which is easily computed during back-propagation. Finally, the evaluation function is shown as:
\begin{equation}\label{eq:te}
    \Theta_\mathrm{TE}(h_i^l)=\frac{1}{T}\sum_{t}\left|\frac{\delta\mathcal{L}(H, h_i^l)}{\delta h_i^l}h_i^l\right|,
\end{equation}
where $h_i^l$ is the activation value of the $i$-th neuron of $l$-th layer and $T$ is the number of the training examples. The criterion is computed on the general-domain data and averaged over $T$. Finally, we prune a certain percentage of neurons and relevant parameters in each target layer based on this criterion.

\textbf{Weight Pruning} We adopt the magnitude-based weight pruning scheme~\cite{SeeLM16}, where the absolute value of each parameter in the target matrix is treated as the importance:
\begin{equation}
    \Theta_\mathrm{AV}(\mathrm{w}_{mn})=|\mathrm{w}_{mn}|, \mathrm{w}_{mn} \in \mathbf{W},
\end{equation}
where $\mathrm{w}_{mn}$ denotes the $m$-th row and $n$-th column parameter of the weight matrix $\mathbf{W}$. The weight matrix $\mathbf{W}$ represents different parts of the model, e.g., embedding layer, attention layer, output layer, etc. Finally, a certain percentage of parameters in each target parameter matrix are pruned.

\subsection{Knowledge Distillation}
Though only limited degradation will be brought in performance after removing the unimportant neurons or parameters, we want to further reduce this loss. To achieve this, we minimize the difference in the output distribution of the unpruned and pruned model. In this work, the general-domain model (parameters denoted as $\theta_G^*$) acts as the teacher model and the pruned model (parameters denoted as $\theta_G$) acts as the student model. So, the objective in this training phase is:
\begin{equation}
\begin{split}
    \mathcal{L}_{\mathrm{KD}}(\theta_G, \theta_G^*)& =  -\frac{1}{K}  \sum_{k=1}^{K} q(\mathbf{y}_k | \mathbf{y}_{<k}, \mathbf{x}; \theta_G^{*}) \\
     & \times \log p(\mathbf{y}_k | \mathbf{y}_{<k}, \mathbf{x}; \theta_G).
\end{split}
\end{equation}
Considering that the general-domain data is not always available in some scenarios when adapting the model to new domains, e.g., continual learning, we adopt the word-level knowledge distillation method using the \textbf{in-domain data}. Because the teacher model is trained on general-domain, it can still transfer the general-domain knowledge to the student model even with the in-domain data. We can fine-tune the pruned model on general-domain if the data is available which can simplify the training procedure. We have also tried the sentence-level knowledge distillation method, but the results are much worse.
The parameters of the teacher model keep fixed during this training phase and the parameters of the pruned model are updated with this KD loss. After convergence, the parameters of the pruned model ($\theta_G$) will be solely responsible for the general-domain and will also participate in the translation of in-domain data. 
These parameters will be kept fixed during the following training phase, so our model won't suffer catastrophic forgetting on the general-domain during the fine-tuning process. 

\subsection{Model Expansion}
After getting the well-trained pruned model, we add new parameters (denoted as $\theta_I$) to it, which expands the model to its original size. Then we fine-tune these newly added parameters with in-domain data, which is supervised by the ground truth sequences. As we have indicated above, the parameters of the pruned model (denoted as $\theta_G$), which are responsible for generating the general-domain translation, keep fixed during this training phase. The objective function is:
\begin{equation}
    \mathcal{L}(\theta_G, \theta_I) = -\frac{1}{K} \sum_{k=1}^{K} \log p(y_k^{*} | \mathbf{y}_{<k}, \mathbf{x}; \theta_G, \theta_I).
\end{equation}
After convergence, the parameters of the pruned model ($\theta_G$) and new parameters ($\theta_I$) are combined together for generating the in-domain translation.

\section{Experiments}
\subsection{Data Preparation}

\textbf{Chinese$\rightarrow$English}. For this task, the general-domain data is from WMT 2017 Zh-En translation task that contains 23.97M sentence pairs. The data is mainly related to the \textbf {News} domain. The newsdev2017 and newstest2017 are chosen as the development and test set, respectively. We choose the parallel sentences with the domain label \textbf {Thesis} from the UM-Corpus~\cite{TianWCQOY14} as our in-domain data. This portion covers 15 journal topics in the research area. We filter out the duplicate sentences and then choose 75K, 1K, and 1K sentences randomly as our training, development, and test data, respectively. We tokenize and truecase the English sentences with Moses scripts.\footnote{http://www.statmt.org/moses/} For the Chinese data, we perform word segmentation by using Stanford Segmenter.\footnote{https://nlp.stanford.edu/} 

\textbf{English$\rightarrow$French}. For this task, the general-domain data is from the UN corpus of the WMT 2014 En-Fr translation task that contains 12.78M sentence pairs, which are mainly related to the \textbf{News} domain. We choose newstest2013 and newstest2014 as our development and test set, respectively. The in-domain data with 53K sentence pairs are from WMT 2019 biomedical translation task, and it is mainly related to the \textbf{Biomedical} domain. We choose 1K and 1K sentences randomly from the corpora as our development and test data, respectively. We tokenize and truecase the corpora.

\textbf{English$\rightarrow$German}. For this task, general-domain data is from the WMT16 En-De translation task which is mainly \textbf {News} texts. It contains about 4.5M sentence pairs. We choose the newstest2013 for validation and newstest2014 for test. For the in-domain data, we use the parallel training data from the IWSLT 2015 which is mainly from the \textbf {Spoken} domain. It contains about 194K sentences. We choose the 2014test for validation and the 2015test for test. We tokenize and truecase the corpora.

Besides, integrating operations of 32K, 32K, and 30K are performed to learn BPE~\cite{SennrichHB16a} on the general-domain data and then applied to both the general-domain and in-domain data. Then we filter out the sentences which are longer than 128 sub-words. For the Zh-En translation task, 44K size of the Chinese dictionary and 33K size of the English dictionary are built based on the general-domain data. For the En-Fr and En-De tasks, 32K size of the dictionaries for the source and target languages are also built on the corresponding general-domain data. 

\begin{table*}[t]
\centering
\resizebox{2.1\columnwidth}!{
\begin{tabular}{c|c|c|cccc|cccc|cccc}
\hline
\multirow{2}{*}{ID} & \multicolumn{2}{c|}{\multirow{2}{*}{System}} & \multicolumn{4}{c|}{Chinese-English} & \multicolumn{4}{c|}{English-French} & \multicolumn{4}{c}{English-German} \\ \cline{4-15} 
 & \multicolumn{2}{c|}{} & \#Para. & Gen. & In. & Avg. & \#Para. & Gen. & In. & Avg. & \#Para. & Gen. & In. & Avg. \\ \hline \hline
0 & \multicolumn{2}{c|}{General} & \multirow{6}{*}{100.5M} & 23.26 & 9.97 & 16.62 & \multirow{6}{*}{93.3M} & \textbf{33.05} & 25.25 & 29.15 & \multirow{6}{*}{94.4M} & 26.22 & 29.53 & 27.88 \\
1 & \multicolumn{2}{c|}{In} &  & 0.87 & 3.64 & $2.26^{-14.36}$ &  & 3.88 & 7.87 & $5.88^{-23.27}$ &  & 12.63 & 24.77 & $18.70^{-9.18}$ \\
2 & \multicolumn{2}{c|}{Fine-tuning} &  & 13.78 & \textbf{17.05} & $15.42^{-1.20}$ &  & 14.03 & \textbf{33.24} & $23.64^{-5.51}$ &  & 23.37 & 32.63 & $28.00^{+0.12}$ \\
3 & \multicolumn{2}{c|}{SeqKD} &  & 14.02 & 13.45 & $13.74^{-2.88}$ &  & 20.37 & 27.93 & $24.15^{-5.00}$ &  & 24.50 & 30.20 & $27.35^{-0.53}$ \\
4 & \multicolumn{2}{c|}{MOL} &  & 18.44 & 13.86 & $16.15^{-0.47}$ &  & 26.58 & 28.05 & $27.32^{-1.83}$ &  & 25.42 & 32.09 & $28.76^{+0.88}$ \\
5 & \multicolumn{2}{c|}{EWC} &  & 17.45 & 15.72 & $16.59^{-0.03}$ &  & 25.29 & 32.24 & $28.77^{-0.38}$ &  & 25.62 & 32.17 & $28.90^{+1.02}$ \\ \hline
6 & \multicolumn{2}{c|}{Full Bias} & 100.6M & 23.26 & 11.55 & $17.41^{+0.79}$ & 93.4M & \textbf{33.05} & 26.47 & $9.76^{+0.61}$ & 94.5M & 26.22 & 30.00 & $28.11^{+0.23}$\\
7 & \multicolumn{2}{c|}{Adapter} & 101.3M & 23.26 & 15.82 & $19.54^{+2.92}$ & 94.1M & \textbf{33.05} & 29.61 & $31.33^{+2.18}$ & 95.3M & 26.22 & 31.54 & $28.88^{+1.00}$ \\
8 & \multicolumn{2}{c|}{MLL} & 117.4M & 23.30 & 16.04 & $19.67^{+3.05}$ & 109.7M & 32.70 & 30.78 & $31.74^{+2.59}$ & 111.2M & 26.08 & 32.07 & $29.08^{+1.20}$ \\ \hline
9 & \multirow{6}{*}{\begin{tabular}[c]{@{}c@{}} PTE  \end{tabular}} & 0+\textbf{NP} & 81.1M & 18.35 & 9.70 & $14.03^{-2.45}$ & 76.3M & 29.27 & 24.86 & $27.07^{-2.60}$ & 77.2M & 24.99 & 26.23 & $25.61^{-2.27}$ \\
10 &  & 9+KD & 81.1M & 22.62 & 9.99 & $16.31^{-0.17}$ & 76.3M & 32.77 & 25.83 & $29.30^{-0.37}$ & 77.2M & 26.04 & 27.71 & $26.88^{-1.00}$ \\
11 &  & 10+FT & 100.5M & 22.62 & 15.94 &  $19.28^{+2.81}$ & 93.3M & 32.77 & 30.69 & $31.73^{+2.07}$ & 94.4M & 26.04 & 32.57* & $29.31^{+1.43}$ \\ \cdashline{3-15}
12 & & 0+\textbf{WP} & 70.4M & 20.74 & 9.54 & $15.14^{-1.48}$ & 65.3M & 29.65 & 25.03 & $27.34^{-1.81}$ & 66.1M & 25.02 & 26.66 & $25.84^{-2.04}$ \\
13 & & 12+KD & 70.4M & 23.50 & 9.77 & $16.64^{+0.02}$ & 65.3M & 32.64 & 25.98 & $29.31^{+0.16}$ &  66.1M & 26.38 & 26.74 & $26.56^{-1.32}$\\
14 & & 13+FT & 100.5M & \textbf{23.50} & \textbf{16.98**} & $\textbf{20.24}^{\textbf{+3.62}}$ & 93.3M & 32.64 & \textbf{33.16**} & $\textbf{32.90}^{\textbf{+3.75}}$ &  94.4M & \textbf{26.38} & \textbf{33.02**} & $\textbf{29.70}^{\textbf{+1.82}}$\\
\hline
\end{tabular}
}
\caption{BLEU scores on three translation tasks. '\#Para.' denotes the number of parameters of the whole model, 'Gen.' and 'In.' denote the BLEU on general-domain and in-domain, and 'Avg.' denotes the average BLEU of the two test sets.
'NP', 'WP', 'KD', and 'FT' represent neuron pruning, weight pruning, knowledge distillation, and fine-tuning, respectively. The numbers on the right of 'PTE' denote that this training phase is based on the previous corresponding models. After knowledge distillation, the parameters in the pruned model (system 10, 13) are fixed, so the general-domain BLEU is unchanged after fine-tuning (system 11, 14). * and ** mean the improvements over the MLL method is statistically significant ($\rho < 0.05$ and $\rho < 0.01$, respectively).~\cite{CollinsKK05}}
\label{tab-mainres}
\end{table*}

\subsection{Systems}
We use the open-source toolkit called {\em Fairseq-py}~\cite{ott2019fairseq} released by Facebook as our Transformer system. The contrast methods can be divided into two categories. The models of the first category are capacity-fixed while the second category are capacity-increased. The first category includes the following systems: 

\noindent \textbullet \ \textbf{General} This baseline system is trained only with the general-domain training data. 

\noindent \textbullet \ \textbf{In} This baseline system is trained only with the in-domain training data. 


\noindent \textbullet \ \textbf{Fine-tuning}~\cite{luong2015stanford} This method just continues to train the general-domain model with the in-domain data.

\noindent \textbullet \ \textbf{SeqKD}~\cite{KimR16} The in-domain source sentences are first translated by the general-domain model. Then the model is further trained with the combined pseudo and real data.

\noindent \textbullet \ \textbf{Multi-objective Learning (MOL)}~\cite{dakwale2017fine} This method is based on the Fine-tuning method. Besides minimizing the loss between the ground truth words and the output distribution of the network, this method also minimizes the cross-entropy between the output distribution of the general-domain model and the network. The final objective is:
\begin{equation}
    \mathcal{L}_{\mathrm{MOL}}(\theta) = \mathcal{L}(\theta) + \alpha \mathcal{L}_{\mathrm{KD}}(\theta)
\end{equation}
where $\alpha$ is the hyper-parameter which controls the contribution of the two parts. The bigger the value, the less degradation on the general-domain. 

\noindent \textbullet \ \textbf{Elastic Weight Consolidation (EWC)}~\cite{ThompsonGKDK19} This method models the importance of the parameters with Fisher information matrix and puts more constrains on the important parameters to let them stay close to the original values during the fine-tuning process. The training objective is:
\begin{equation}
    \mathcal{L}_{\mathrm{EWC}}(\theta) = \mathcal{L}(\theta) + \alpha \sum_{i} F_i (\theta_i - \theta_i^G)^2
\end{equation}
where $i$ represents the $i$-th parameter and $F_i$ is the modeled importance for the $i$-th parameter.

The second category indcludes the following three systems:

\noindent \textbullet \ \textbf{Full Bias}~\cite{MichelN18} This method adds domain-specific bias term to the output softmax layer and only updates the term as other parts of the general-domain model keep fixed.  

\noindent \textbullet \ \textbf{Adapter}~\cite{BapnaF19} This methods injects domain-specific adapter modules into each layer of the general-domain model. Each adapter contains a normalization layer and two linear projection layers. The adapter size is set to 64. 

\noindent \textbullet \ \textbf{Multiple-output Layer Learning (MLL)}~\cite{dakwale2017fine} The method modifies the general-domain model by adding domain-specific output layer for the in-domain and learning these domain specific parameters with respective learning objective. The training objective is:
\begin{equation}
    \mathcal{L}_{\mathrm{MLL}}(\theta_S, \theta_G, \theta_I) = \mathcal{L}(\theta_S, \theta_I) + \alpha \mathcal{L}_{\mathrm{KD}}(\theta_S, \theta_G)
\end{equation}
where $\theta_S$ is the domain-shared parameters, $\theta_G$ and $\theta_I$ denote the domain specific parameters for the general-domain and in-domain, respectively.

\noindent \textbullet \ \textbf{Our Method - Pruning Then Expanding (PTE)} Our model is trained just as the Method section describes. For the neuron pruning scheme, we prune the last 10\% unimportant neurons; for the weight pruning scheme, we prune the last 30\% unimportant parameters. To better show the ability of our method, we report the general- and in-domain performance after each training phase.
 
\noindent \textbf{Implementation Details} All the systems are implemented as the base model configuration in~\citet{VaswaniSPUJGKP17} strictly. We set the hyper-parameter $\alpha$ to $1$ for MOL, EWC, and MLL and we will do more analysis on the impact of this hyper-parameter in the next section. 
We set the learning rate during fine-tuning process to $7.5\times10^{-5}$ for all the systems after having tried different values from $1.5\times10^{-6}$ to $1.5\times10^{-3}$.
In both of our methods, we don't prune the layer-normalization layers in the encoder and decoder, which can make training faster and more stable. For the neuron pruning method, we also don't prune the first layer of the encoder and the last layer of the decoder.
Just like the work of~\citet{dakwale2017fine}, the domain of the test data is known in our experiments. Besides, we use beam search with a beam size of 4 during the decoding process.

\begin{figure*}[t]
    \centering
    \subfigure[Zh$\rightarrow$En]{
        \includegraphics[width=0.66\columnwidth]{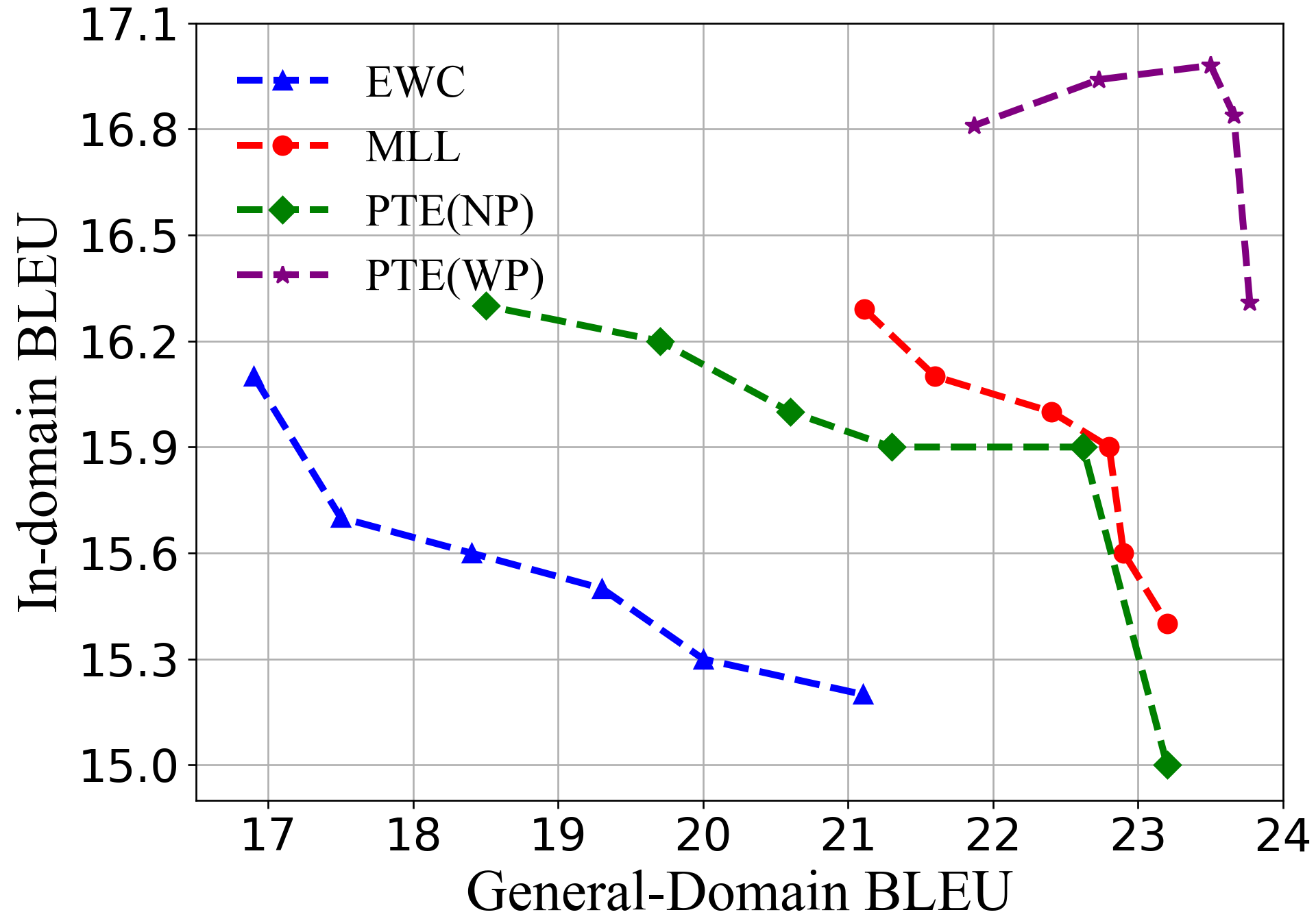}
    }
    \subfigure[En$\rightarrow$Fr]{
        \includegraphics[width=0.66\columnwidth]{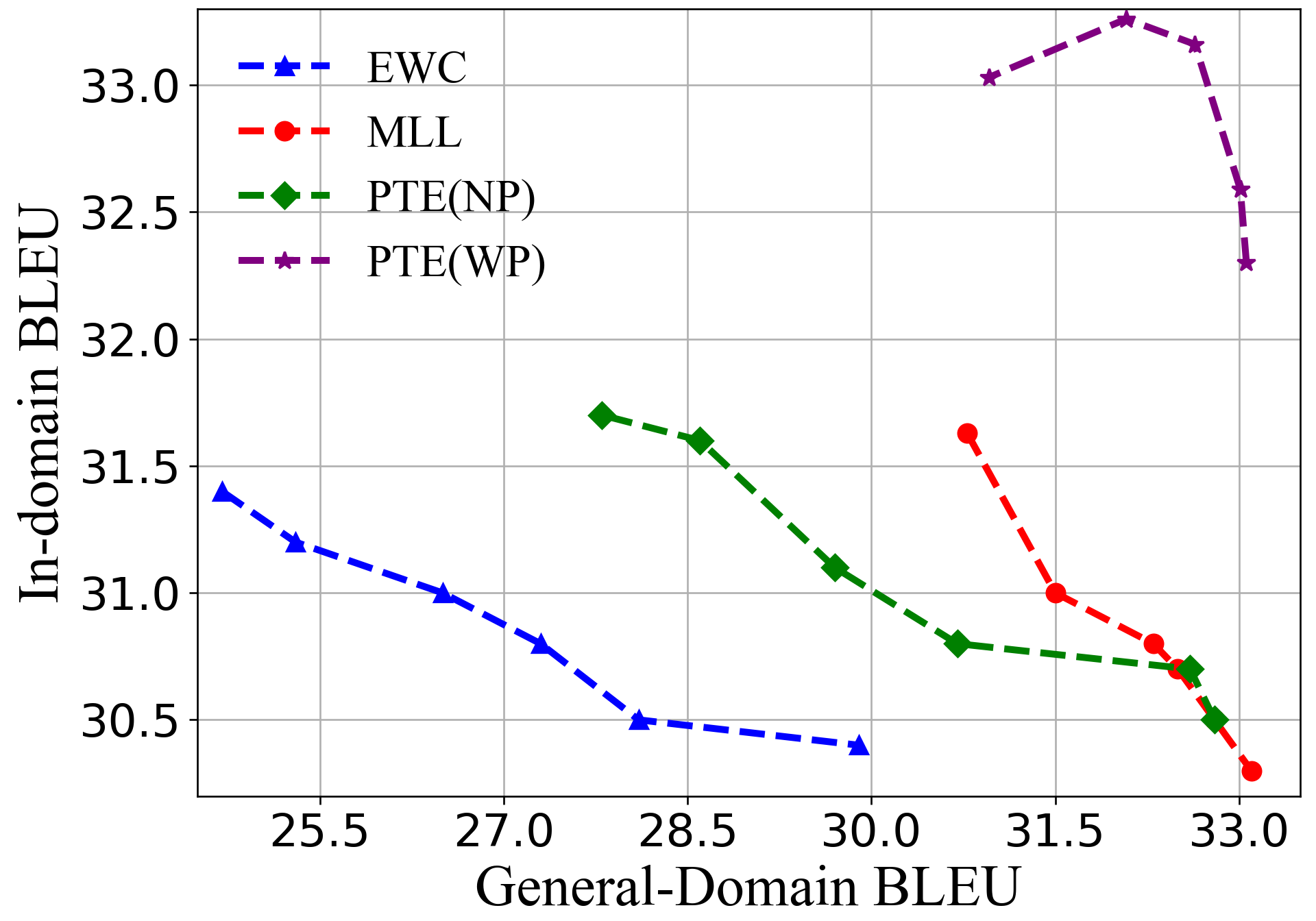}
    }
    \subfigure[En$\rightarrow$De]{
        \includegraphics[width=0.66\columnwidth]{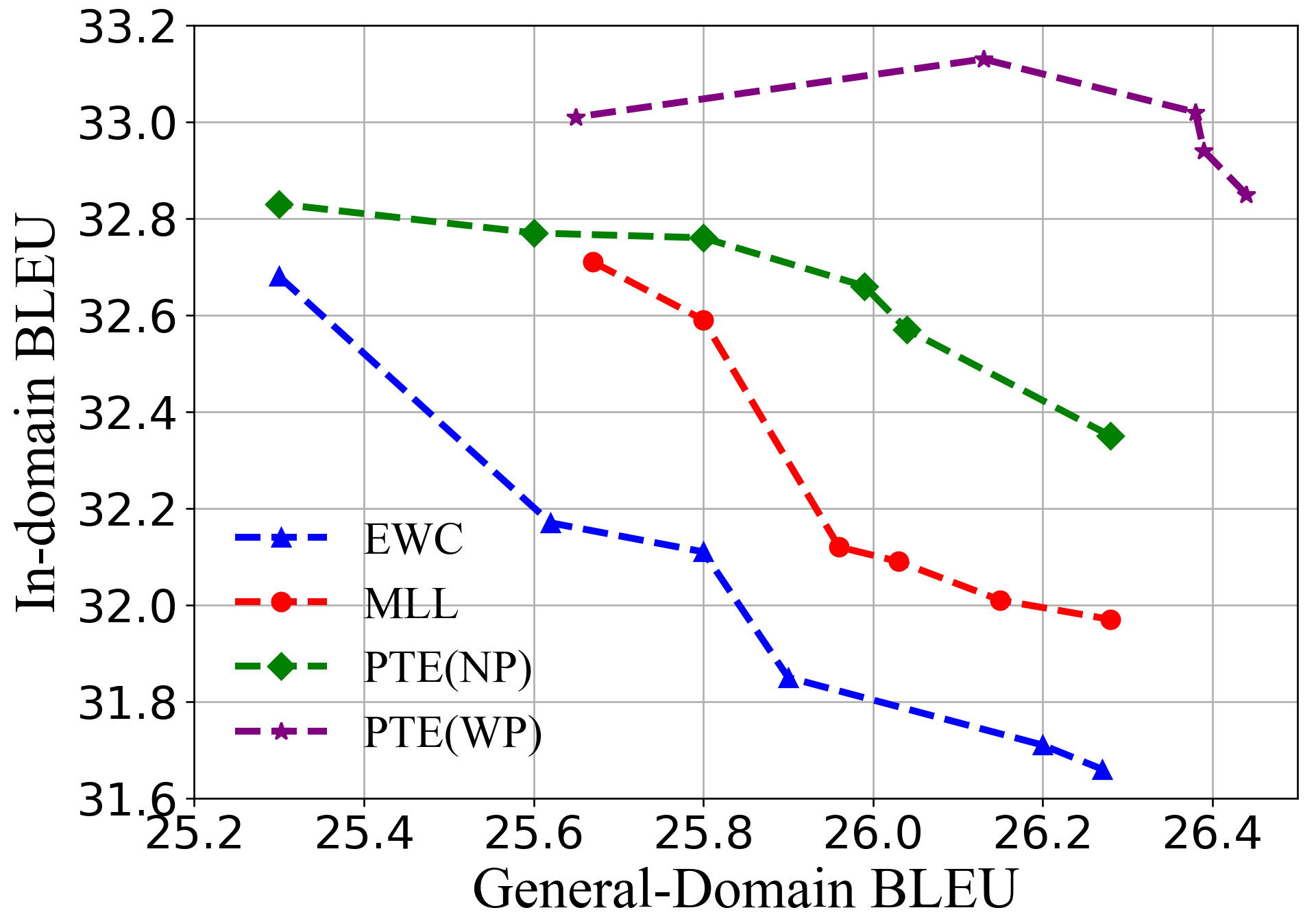}
    }
    \caption{ The performance trade-off with different hyper-parameters. the x-axis is general-domain BLEU and the y-axis is in-domain BLEU. The closer the point is to the upper right corner, the better the performance.}
    \label{fig:hyper}
\end{figure*}

\subsection{Main Results}

The final translation is detokenized and then the quality is evaluated using the $4$-gram case-sensitive BLEU~\cite{PapineniRWZ02} with the {\em SacreBLEU} tool~\cite{post-2018-call}.\footnote{BLEU+case.mixed+numrefs.1+smooth.exp+tok.13a\\+version.1.3.6} The results are given in Table~\ref{tab-mainres}. In all the datasets, our weight pruning method outperforms all the baselines. Furthermore, we get the following conclusions:

First, the contrast capacity-fixed methods can't handle large domain divergence and still suffer catastrophic forgetting. 
They perform well in the En-De translation task, where the data distributions are similar. They can significantly improve the in-domain translation quality without excessive damage to the general-domain translation quality. 
However, they perform worse in the En-Fr and Zh-En translation tasks with more different data distributions. The in-domain data contains many low-frequency or out-of-vocabulary tokens of the general-domain data.
In this situation, these methods either bring limited in-domain improvements or degrade the general-domain performance too much.
In contrast, our method is superior to them in all tasks, especially on the more different domains. This also validates our motivation.

Second, the capacity-increased methods can better deal with domain divergence. Compared with them, our method can achieve larger improvements on in-domain since we actually allocate more parameters for in-domain than the capacity-increased methods.
Besides, our methods are also more convenient to use in practice because we don't need to specialize the model architecture. The pruning ratio is the only hyper-parameter needed tuning. 

Lastly, both of our methods are immune to large domain divergence. Moreover, the knowledge distillation can bring modest improvements on the general domain.
Compared with the neuron pruning method, the weight pruning method is more effective since it can prune and reutilize more parameters with smaller performance degradation.


\section{Analysis}

\begin{table}[t]
\centering
\resizebox{\columnwidth}!{
\begin{tabular}{c|ccccc}
\hline
System & \#Para. & Gen. & T & S & E \\ \hline \hline
Fine-tuning  & 100.5M & 12.58 & \textbf{16.99} & 18.64 & 19.43 \\ 
Adapter & 102.9M & 23.26 & 15.82 & 17.83 & 18.68 \\
MLL & 151.2M & 22.60 & 16.24 & 18.27 & 18.39 \\
PTE(WP) & 100.5M & \textbf{23.78**} & 16.85* & \textbf{18.69} & \textbf{19.55**} \\ \hline  

\end{tabular}
}
\caption{BLEU of different domains on the Zh-En task. 'T', 'S', and 'E' denote the in-domain of Thesis, Spoken, and Education, respectively. 'PTE(WP)' denotes our weight-pruning based method.}
\label{tab:multi}
\end{table}

\subsection{Adapting to Multi-Domain}
We conduct experiments under the multi-domain scenario, which lets the model adapt to several different domains. Except for the training data used in the main experiments of the Zh-En task, which are related to the \textbf{News} and \textbf{Thesis} domain, we add two datasets from other domains, namely, \textbf{Spoken} and \textbf{Education}. Both of them are chosen randomly from the UM-corpus. Each of them contains about 75K, 1K, and 1K sentence pairs in the training, development, and test set. We test our weight-pruning based method and still prune last 30\% unimportant parameters. We compare our method with the basic fine-tuning system and more effective capacity-increased method. 
The results are shown in Table~\ref{tab:multi}. It shows that our method can get significant improvements on all the domains.

\subsection{Effects of Different Hyper-parameters}
For the MOL, EWC, and MLL methods, the hyper-parameter $\alpha$ controls the trade-off between the general- and in-domain performance.
As for our method, the proportion of model parameters to be pruned has a similar effect.  
To better show the full general- and in-domain performance trade-off, we conduct experiments with different hyper-parameters. We compare our method with the best capacity-fixed method EWC and best capacity-increased method MLL.
For the EWC and MLL method, we vary $\alpha$ from $0.25$ to $2.5$. We vary the pruning proportion from $5\%$ to $30\%$ for our neuron-pruning method and from $10\%$ to $50\%$ for our weight-pruning method. The results are shown in Figure~\ref{fig:hyper}. It shows that our method outperforms EWC at all the operating points significantly. Besides, our neuron-pruning method can achieve comparable results as MLL and our weight-pruning method can surpass it with fewer parameters.

\begin{table}[t]
\centering
\resizebox{\columnwidth}!{
\begin{tabular}{c|c|c|ccc}
\hline
ID & \multicolumn{2}{c}{System} & Gen. & In. & Avg. \\ \hline \hline
0 & \multicolumn{2}{c|}{General} & 23.26 & 9.97 & 16.62 \\ \hline
12 & \multirow{3}{*}{\begin{tabular}[c]{@{}l@{}}PTE\end{tabular}} & 0+WP & 20.74 & 9.54 & $15.14^{-1.48}$ \\
13 &  & 12+KD & 23.50 & 9.77 & $16.64^{+0.02}$ \\
14 &  & 13+FT & 23.50 & 16.98 & $20.24^{+3.62}$ \\ \hline
15 & \multirow{3}{*}{Random} & 0+WP & 12.47 & 5.18 & $8.83^{-7.79}$\\
16 &  & 15+KD & 14.69 & 5.48 & $10.09^{-6.53}$\\
17 &  & 16+FT & 14.69 & 16.03 & $15.36^{-1.26}$ \\ \hline
18 & Selective FT & 0+FT & 13.74 & 16.58 & $15.16^{-1.46}$\\ \hline 
\end{tabular}
}
\caption{Results of the ablation study. 'Random' denotes the parameters are randomly pruned. 'Selective FT' denotes only the unimportant parameters are fine-tuned. Other denotations are the same as in Table~\ref{tab-mainres}.}
\label{tab:random}
\end{table} 
\subsection{Ablation Study}
To further understand the impact of each step of our method, we perform further studies by removing or replacing certain steps of our method. We first investigate the necessity of parameter importance evaluation. We train another three models following our method but with the parameters randomly pruned. The results are given in Table~\ref{tab:random}. It shows that random pruning will give excessive damage to general-domain. Besides, we also train another model that skips the model pruning and knowledge distillation steps and directly fine-tune the unimportant parameters. At last, we perform translation with the whole model on both the general- and in-domain. The results show that the change of unimportant parameters will also lead to catastrophic forgetting on general-domain, which shows the necessity of ``divide and conquer''.

\subsection{Effects of Data Distribution Divergence}
\begin{figure}[t]
    \centering
    \includegraphics[width=\columnwidth]{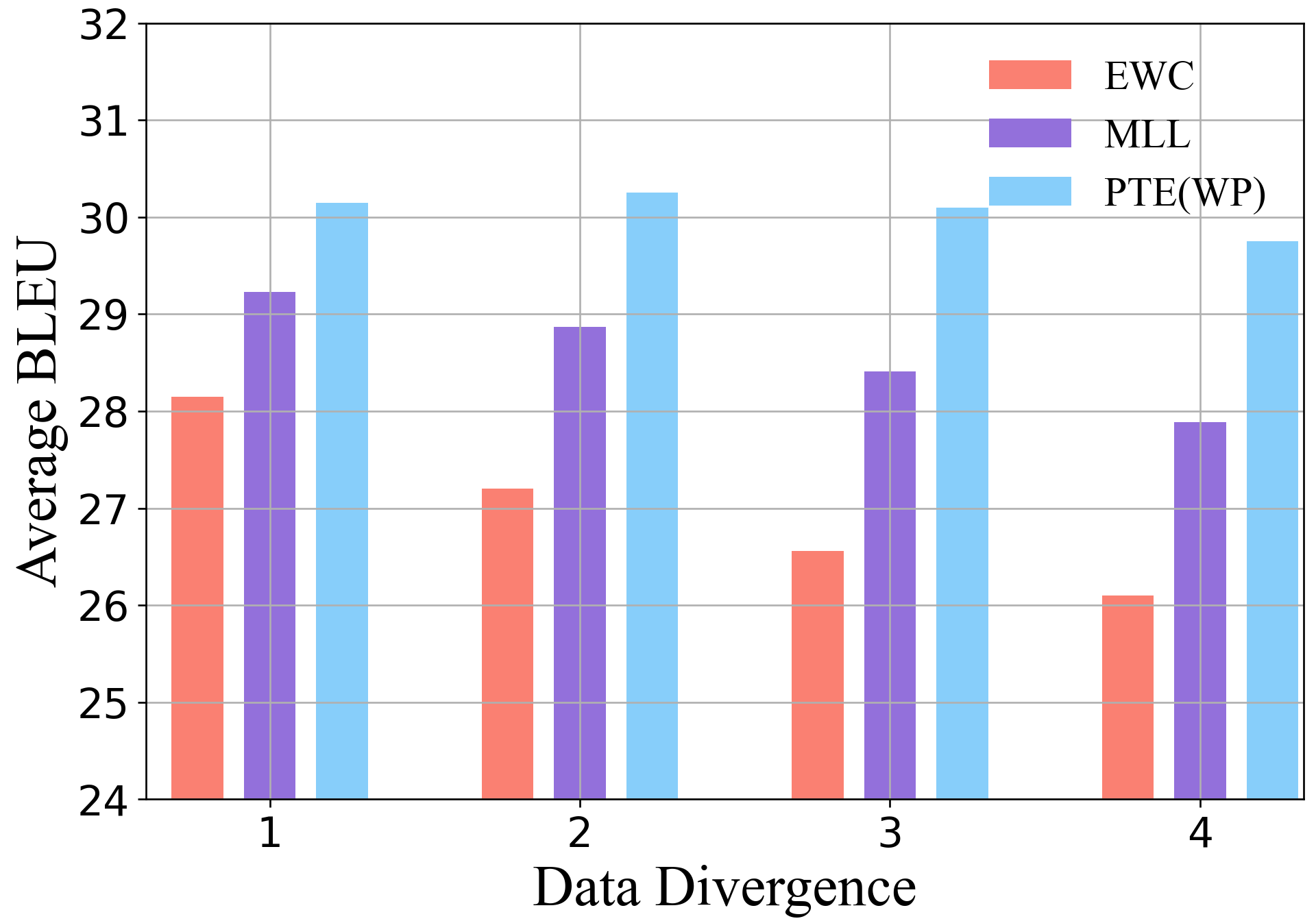}
    \caption{The average BLEU with different domain divergences on the En$\rightarrow$Fr translation task. The x-axis represents the divergence of the data distribution and the larger the value is, the general- and in-domain data are more different.}
    \label{fig:4}
\end{figure}

To further prove that our method is better at dealing with large domain divergence, we conduct experiments on the En-Fr translation task. Following the method in~\citet{MooreL10}, we score and rank each in-domain sentence pair by calculating the per-word cross-entropy difference between the general- and in-domain language model:
\begin{equation}
    \mathtt{Score} = (H_G(s)-H_I(s))+(H_G(t)-H_I(t))
\end{equation}
where $H$ denotes the language model which is trained with Srilm~\cite{Stolcke02}, $s$ and $t$ denote the source and target sentence. 
Then, we split the in-domain data into four parts with equal size and train new models with them separately. We compare our weight pruning based method with the EWC and MLL methods.
The results are shown in Figure~\ref{fig:4}. It shows that we can get larger improvements as the data divergence gets larger. 

\section{Related Work}
\textbf{Domain Adaptation} Recent work on DA can be divided into two categories according to the use of training data. The first category, which is also referred to as multi-domain adaptation, needs the training data from all of the domains. \citet{chu2017empirical} fine-tunes the model with the mix of the general-domain data and over-sampled in-domain data. \citet{KobusCS17} adds domain-specific tags to each sentence. \citet{ZhangSKMCD19} applies curriculum learning to the DA problem. 
\citet{BritzLP17} adds a discriminator to extract common features across domains. There are also some work~\cite{ZengSWLXYZ18,ZengLSGLYL19,GuFL19} that adds domain-specific modules to the model to preserve the domain-specific features. 
\citet{currey2020distilling} distills multiple expert models into a single student model. 
The work of~\citet{liang2020finding} has a similar motivation with ours which also fix the important parameters and prune the unimportant parameters. Compared with their method, our method doesn't need to store the general-domain training data and our method has less degradation on general-domain because we adopt the knowledge distillation method. 


The second category, which is also referred to as continual learning, only needs the data from the new domain and the model in use. 
The biggest challenge for this kind of work is the catastrophic forgetting. 
\citet{luong2015stanford} fine-tunes the general-domain model with the in-domain data. \citet{FreitagA16} ensembles the general-domain model and the fine-tuned model for generating. \citet{SaundersSGB19} investigates adaptive ensemble weighting for inference.  \citet{KhayrallahTDK18} and \citet{ThompsonGKDK19} add regularization terms to let the model parameters stay close to their original values. \citet{dakwale2017fine} minimizes the cross-entropy between the output distribution of the general-domain model and the fine-tuned model. \citet{MichelN18} adds domain-specific softmax bias term to the output layer. \citet{BapnaF19} injects domain-specific adapter modules into each layer of the general-domain model. \citet{WuebkerSD18} only saves the domain-specific offset based on the general-domain model. \citet{WangMPC20} achieves efficient lifelong learning by establishing complementary learning systems. \citet{SatoS0TK20} adapts the vocabulary of a pre-trained NMT model to the target domain. 

Overall, our work is related to the second type of approach, which is more flexible and convenient in practice. 
The work of~\citet{ThompsonGKDK19} and \citet{dakwale2017fine} are most related to our work. Compared with~\citet{ThompsonGKDK19}, our work is better at dealing with large domain divergence, since we add domain-specific parts to the model. In contrast to~\citet{dakwale2017fine}, our model divides each layer of the model into domain-shared and domain-specific parts, which increases the depth of the in-domain model, intuitively. Besides, our method doesn't need to add parameters, but it can be easily extended when necessary. 

\textbf{Model Pruning} Model pruning usually aims to reduce the model size or improve the inference efficiency. \citet{SeeLM16} examines three magnitude-based pruning schemes. \citet{ZhuG18} demonstrates that large-sparse models outperform comparably-sized small-dense models. \citet{WangWLT20} improves the utilization efficiency of parameters by introducing a rejuvenation approach. \citet{Lan2020ALBERT} presents two parameter reduction techniques to lower memory consumption and increase the training speed of BERT. 

\section{Conclusion}
In this work, we propose a domain adaptation method based on the importance of neurons and parameters of the NMT model. We make the important ones compromise between domains while unimportant ones focus on in-domain. Based on this, our method consists of several steps, namely, model pruning, knowledge distillation, model expansion, and fine-tuning.
The experimental results on different languages and domains prove that our method can achieve significant improvements with model capacity fixed. 
Further experiments prove that our method can also improve the overall performance under the multi-domain scenario.

\section*{Acknowledgements}
We thank all the anonymous reviewers for their insightful and valuable comments. 

\bibliography{emnlp2020}
\bibliographystyle{acl_natbib}

\end{document}